\documentclass[english]{article}
\usepackage[utf8]{inputenc}
\usepackage[T1]{fontenc}
\usepackage{babel}
\usepackage{amsmath}
\usepackage{csquotes}
\usepackage{amstext}
\usepackage{amssymb}
\usepackage{graphicx}
\usepackage{subcaption}
\usepackage{fancyhdr}
\usepackage{siunitx}
\usepackage{comment}
\usepackage{graphicx}

\usepackage{biblatex}
\begin{document}

\title{Reinforcement learning using Deep $Q$ networks and $Q$ learning accurately localizes brain tumors on MRI with very small training sets}

\author{ \textbf{Joseph Stember}$^1$
\and 
\textbf{Hrithwik Shalu}$^2$}
\maketitle
\thispagestyle{fancy}
\noindent
\textsuperscript{1}Memorial Sloan Kettering Cancer Center, New York, NY, US, 10065 
\\
\textsuperscript{2}Indian Institute of Technology, Madras, Chennai, India, 600036
\\
\noindent
\textsuperscript{1}joestember@gmail.com
\\
\textsuperscript{2}lucasprimesaiyan@gmail.com 
\\

\begin{abstract}

\textit{Purpose}
Supervised deep learning in radiology suffers from notorious inherent limitations: 1) It requires large, hand-annotated data sets; 2) It is non-generalizable; and 3) It lacks explainability and intuition. We have recently proposed that reinforcement learning addresses all three of these limitations, applying the methodology to images with radiologist eye tracking points, which limits the state-action space. Here, we generalize Deep Q Learning to a gridworld-based environment so that only the images and image masks are required.
\\
\indent \textit{Materials and Methods}
We trained a Deep $Q$ network on 30 two-dimensional image slices from the BraTS brain tumor database. Each image contained one lesion. We then tested the trained Deep Q network on a separate set of 30 testing set images. For comparison, we also trained and tested a keypoint detection supervised deep learning network on the same set of training/testing images.
\\
\indent \textit{Results}
Whereas the supervised approach quickly overfit the training data and predictably performed poorly on the testing set (11\% accuracy), the Deep $Q$ learning approach showed progressive improved generalizability to the testing set over training time, reaching 70\% accuracy.
\\
\indent \textit{Conclusion}
We have shown a proof-of-principle application of reinforcement learning to radiological images using 2D contrast-enhanced MRI brain images with the goal of localizing brain tumors. This represents a generalization of recent work to a gridworld setting naturally suitable for analyzing medical images.

\end{abstract}

\section*{Introduction}

In recent work \cite{stember2020deep}, we introduced the concept of radiological reinforcement learning (RL): the application of reinforcement learning to analyze medical images. We sought to address three key shortcomings in current supervised deep learning approaches:

\begin{enumerate}
  \item Requirement of large amounts of expert-annotated data.
  \item Lack of generalizability, making it “brittle” and subject to grossly incorrect predictions when even a small amount of variation is introduced. This can occur when applying a trained network to images from a new scanner, institution, and/or patient population \cite{wang2020inconsistent,goodfellow2014explaining}. 
  \item Lack of insight or intuition into the algorithm, thus limiting confidence needed for clinical implementation and curtailing potential contributions from non-AI experts with advanced domain knowledge (e.g., radiologists or pathologists) \cite{buhrmester2019analysis,liu2019comparison}.
\end{enumerate}

Our initial proof-of-principle application of RL to medical images used 2D slices of image volumes from the publicly available BraTS primary brain tumor database \cite{menze2014multimodal}. These T1-post-contrast images included one tumor per image. In addition, our images included an overlay of eye tracking gaze points obtained during a previously performed simulated image interpretation. The state-action space was limited to the gaze plots, consisting of the gaze points for that image. The gaze plots were essentially one-dimensional, and the various possible agent states were defined by location along the gazeplot. Actions were defined by the agent moving anterograde versus retrograde along the gaze plot, or by staying still. As a localization task, the goal was for the agent to reach the lesion. Using the manually traced tumor mask images, a reward system was introduced that incentivized finding and staying within the lesion, and discouraged staying still while the agent was still outside the tumor \cite{stember2020deep}. 

The results from this study showed that RL has the potential to make meaningful predictions based on very small data sets (in this case, 70 training set images). Supervised deep learning woefully overfit the training set, with unsurprisingly low accuracy on the testing set (< 10\%). In contrast, RL improved steadily with more training, ultimately predicting testing set image lesion location with over 80\% accuracy \cite{stember2020deep}. 

However, the system studied was not generalized, as it included eye tracking points, which are usually not available with radiological images. Additionally, the eye tracking points confined the state-action space to one dimension. In order to apply RL more generally to medical images, we must be able to analyze raw images along with accompanying image masks without the need for eye tracking gaze plots. 

In this study, we extended the approach to show that RL can effectively localize lesions using a very small training set using the gridworld framework, which requires only raw images and the accompanying lesion masks. This represents an important early step in establishing that RL can effectively train and make predictions about medical images. This can ultimately be extended to 3D image volumes and more sophisticated implementations of RL. Gridworld is a classic, paradigmatic environment in RL \cite{sutton2018reinforcement}. Given their pixelated character, medical images tiled with a gridworld framework provide a natural, readily suitable environment for our implementation.

\section*{Methods}

\subsection{Basic terms}

Following the basic approach of our recent work, we analyzed 2D image slices from the BraTS public brain tumor database \cite{menze2014multimodal}. These slices were randomly selected from among T1-weighted contrast-enhanced image slices that included brain tumor. Images from around the level of the lateral ventricles were selected. 

As in our recent work, we implementated a combination of standard $TD(0)$ $Q$-learning with Deep $Q$ learning (DQN). The key difference was how we defined the environment, states, and actions. 

We divided the image space of the $240 \times 240$ pixel images by grids spaced 60 pixels, so that our agent occupied the position of a $60 \times 60$ pixel block, shown in Figure \ref{fig:fig_1a}. The initial state for training and testing images was chosen to be the top-left block (figure \ref{fig:fig_1a}). The action space consisted of: 1) staying at the same position, 2) moving down by one block, or 3) moving to the right by one block. In other words, introducing some notation, the action space $\mathcal{A} \in \mathbb{N}_{0}^{3}$, consisting of three non-negative integers, is defined by:
\begin{equation} 
\mathcal{A} = \begin{pmatrix} 1 \\ 2 \\ 3 \end{pmatrix} = \begin{pmatrix} \text{stay still} \\ \text{move down} \\ \text{move right} \end{pmatrix} \text{.} \label{action_eqn}
\end{equation}
To each of the possible actions, $a \in A$, given a policy $\pi$, there is corresponding action value depending on the state, $Q^{\pi}(s,a)$, defined by:
\begin{equation}
    Q^{\pi}(s,a)=E_{\pi}\{R_t \arrowvert s_t=s,a_t=a\}=E_{\pi}\{\sum_{k=0}^{\infty} \gamma^k r_{t+k+1} \arrowvert s_t=s,a_t=a \} \text{,}
\end{equation}
where $R_t$ is the total cumulative reward starting at time $t$ and $E_{\pi}\{R_t \arrowvert s_t=s,a_t=a\}$ is the expectation for $R_t$ upon selecting action $a$ in state $s$ and subsequently picking actions according to $\pi$. 

\subsection{Training: sampling and replay memory buffer}

\begin{figure}
\centering
    \begin{subfigure}{.4\textwidth}
        \centering
        \includegraphics[width=\linewidth]{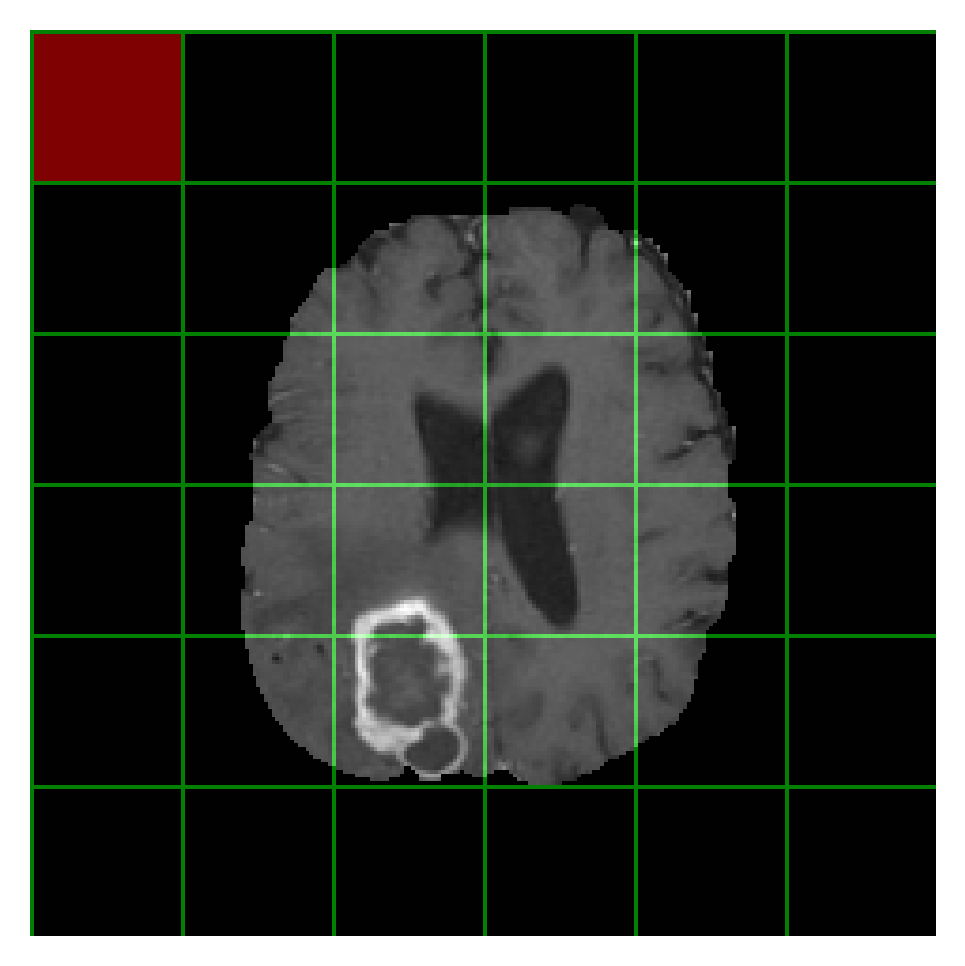}
        \caption{}\label{fig:fig_1a}
    \end{subfigure} %
    \begin{subfigure}{.4\textwidth}
        \centering
        \includegraphics[width=\linewidth]{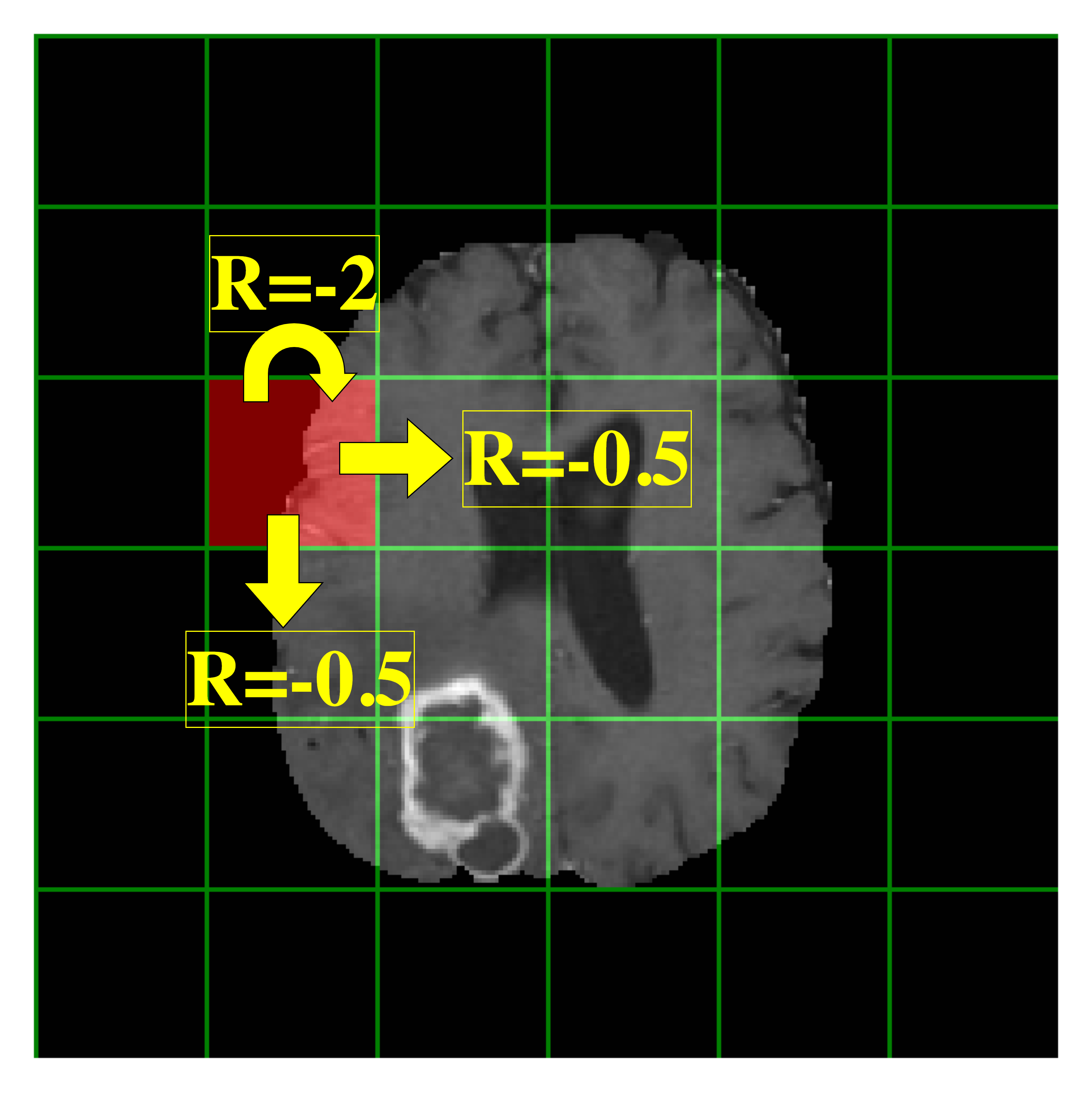}
        \caption{}\label{fig:fig_1b}
    \end{subfigure} %
    \begin{subfigure}{.4\textwidth}
        \centering
        \includegraphics[width=\linewidth]{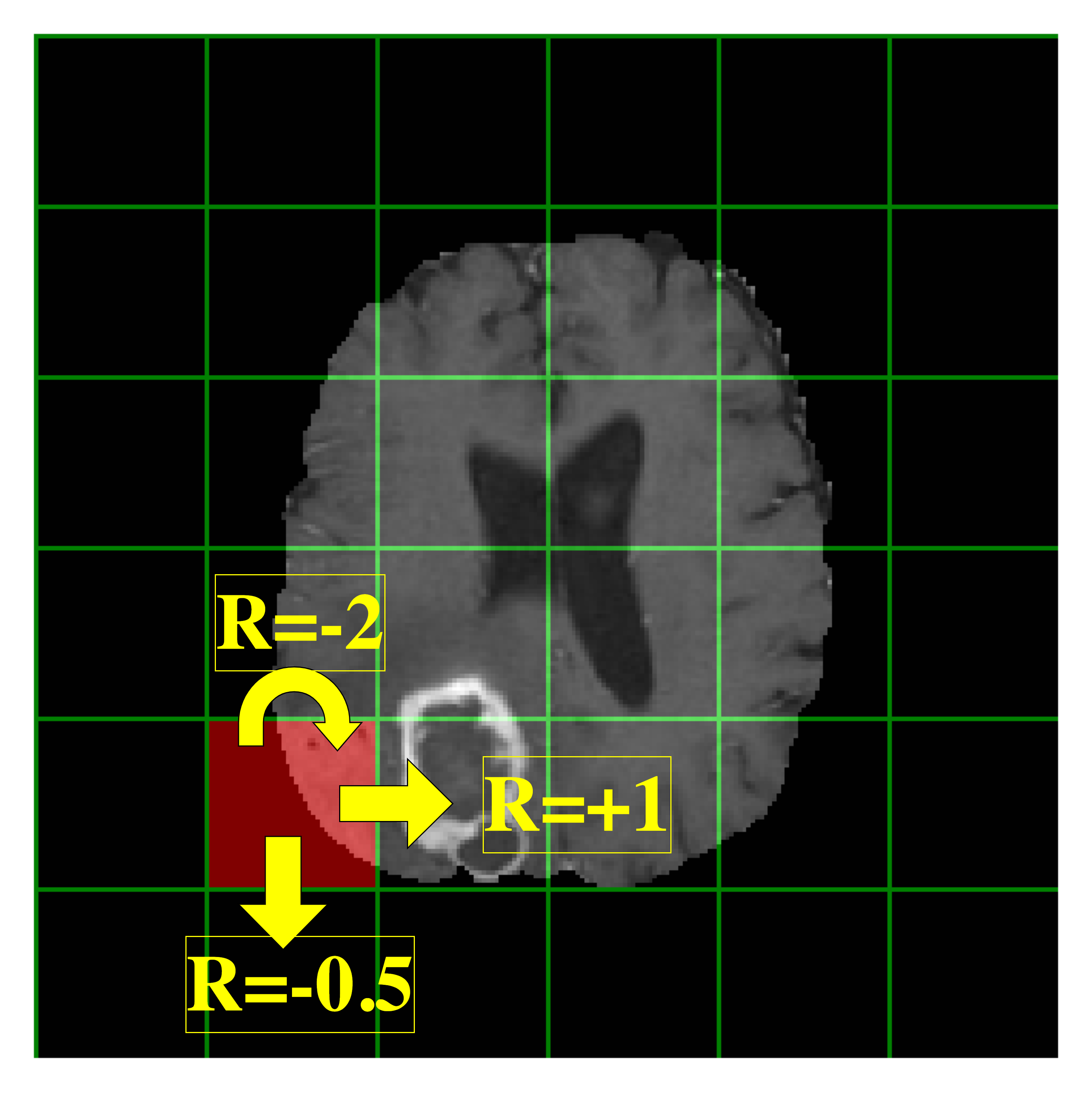}
        \caption{}\label{fig:fig_1c}
    \end{subfigure}
    \begin{subfigure}{.4\textwidth}
        \centering
        \includegraphics[width=\linewidth]{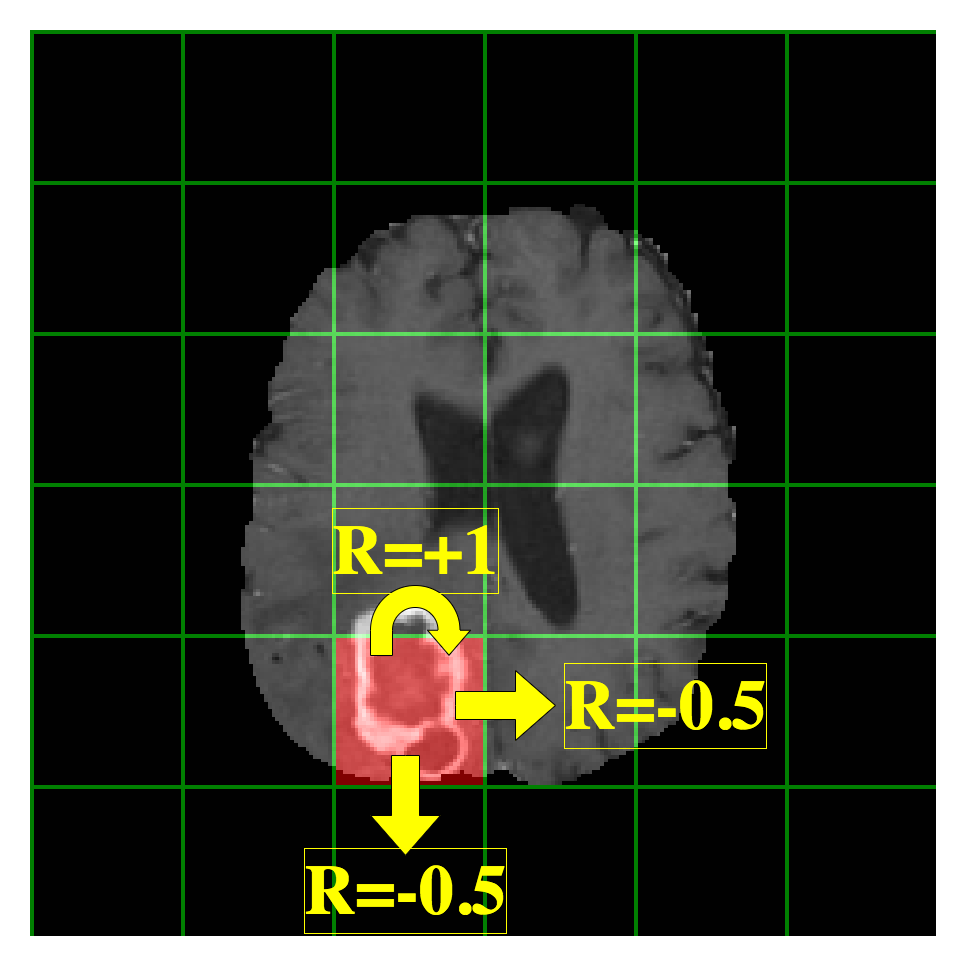}
        \caption{}\label{fig:fig_1d}
    \end{subfigure}
    
\caption{Environment and reward scheme. (a) Shows the initial state for all episodes, $s_1$, with the agent in the upper left corner. (b) Displays the rewards for the three possible actions. When the agent is not in a position overlapping the lesion, staying in place gets the biggest penalty (reward of -2), with a lesser penalty if the agent moves (reward of -0.5). (c) Shows the rewards for the possible actions in the state just to the left of the mass. Moving toward the lesion so that the agent will coincide with it receives the largest possible and only positive reward (+1). (d) Shows the state with the agent coinciding with the lesion. Here we want the agent to stay in place, and thus reward this action with a +1 reward. }
\end{figure}

In each of the $N_{\text{episodes}}=90$ episodes of training, we sample randomly from the $30$ training set images. For each such image we subdivide into grids, and initialize such that the first state $s_1$ is in the upper left block (Figure \ref{fig:fig_1a}). We select each action $a_t$ at time step $t$ according to the off-policy epsilon-greedy algorithm, which seeks to balance exploration of various states with exploiting of known best policy, according to:
\begin{equation}
    a_t =
    \begin{cases}
        max_{a \in A} \{ Q_t ( a ) \} & \text{with probability $\epsilon$} \\
        \text{random action in $A$} & \text{with probability $1-\epsilon$} \text{.}
    \end{cases}
\end{equation}
for the parameter $\epsilon < 1$. We used an initial $\epsilon$ or $0.7$ to allow for adequate exploration. As $Q$ learning proceeds and we wish to increasingly favor exploitation of a better known and more optimal policy, we set $\epsilon$ to decrease by a rate of $1 \times 10^{-4}$ per episode. The decrease continued down to a minimum value $\epsilon_{\text{min}}=1 \times 10^{-4}$, so that some amount of exploring would always take place. 

Our reward scheme is illustrated for sample states in Figures \ref{fig:fig_1b}--\ref{fig:fig_1d}. The reward $r_t$ is given by
\begin{equation}
    r_t = \begin{cases}
        -2 \text{, if agent is outside the lesion and staying still} \\
        +1 \text{, if agent overlaps the lesion and is staying still} \\
        -0.5 \text{, if agent moves to position outside the lesion} \\
        +1 \text{, if agent moves to position overlapping the lesion} \text{.}
    \end{cases}
\end{equation}

\subsection{Replay memory buffer}

In general, for each time $t$ we thus have state $s_t$, the action we have taken $a_t$, for which we have received a reward $r_t$ and which brings our agent to the new state $s_{t+1}$. We store these values in a tuple, called a transition, as $\mathcal{T}_t=(s_t,a_t,r_t,s_{t+1})$. For each successive time step, we can stack successive transitions as rows in a transition matrix $\mathbb{T}$. We do so up to a maximum size of $N_{\text{memory}}=15,000$ rows. These represent the replay memory buffer, which allows the CNN that predicts $Q$ values to sample and learn from past experience sampling from the environments of the various training images. Then, we use $\mathbb{T}$ to train the CNN and perform $Q$ learning. The value of $N_{\text{memory}}=15,000$ was chosen to be as large as possible without overwhelming the available RAM. 

\subsection{Training: Deep $Q$ network and $Q$ learning}

\begin{figure}
\centering
\advance\leftskip-3cm
\includegraphics[width=18cm,height=10cm]{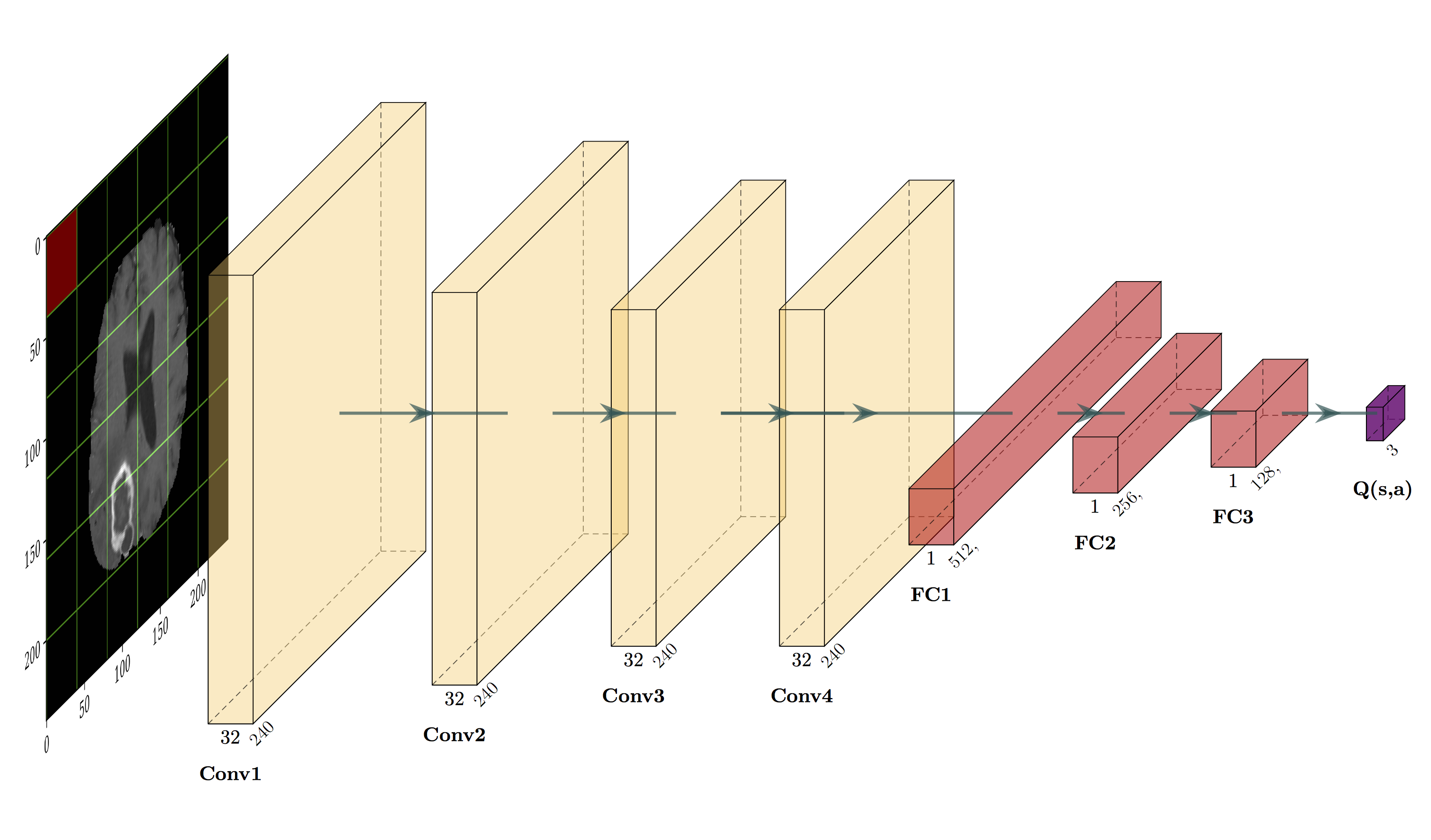}
\caption{Deep Q network architecture. Of note, the output consists of the three $Q$ values corresponding to the three possible actions. A sample input image representing the initial state $s_1$ is also noted. }
\label{fig:dqn_architecture}
\end{figure}

Using a CNN to approximate the function $Q_t(a)$, we give the CNN the name of Deep $Q$ network (DQN). The architecture of the DQN, shown in Figure \ref{fig:dqn_architecture}, is very similar to that of our recent work. It takes the state as input, using $3 \times 3$ kernels with stride of $2$ and padding such that the resulting filter sizes are unchanged. We produced $32$ filters at each convolution operation. The network consisted of four such convolutional filters in sequence, using exponential linear unit (elu) activation. The last convolution layer was followed by a $512$-node layer with elu activation, followed by a few more fully connected layers and ultimately to a $3$-node output layer representing the $3$ actions and corresponding $Q$ values. 

Our DQN loss is the difference between the $Q$ values resulting from a forward pass of the DQN, which at time step $t$ we shall denote as $Q_{\text{DQN}}$, and the “target” $Q$ value, $Q_{\text{target}}$, computed by the Bellman equation. The latter updates by sampling from the environment and experiencing rewards. Denoting the forward pass by $F_{\text{DQN}}$, we can obtain state-action values by
\begin{equation}
    Q_{\text{DQN}}^{(t)}=F_{\text{DQN}}(s_t) \text{.}
\end{equation}
But the function approximation we wish to learn is for the \textit{optimal} state-action values, which maximize expected total cumulative reward. We do so by $Q$ learning, in which the model learns from sampled experience, namely individual state-action pair-associated rewards. The method we used in our prior study, and employ here, is temporal difference in its simplest form: $TD(0)$. With $TD(0)$ the state-action value function is updated in each step of sampling to compute the $TD(0)$ target, denoted by $Q_{\text{target}}^{(t)}$:
\begin{equation}
    Q_{\text{target}}^{(t)} = r_t + \gamma max_a Q(s_{t+1},a) \text{,}
\end{equation}
where $\gamma$ is the discount factor and $max_a Q(s_{t+1},a)$ is another way of writing the state value function $V(s_{t+1})$. The key part of the environment sampled is the reward value $r_t$. Over time, with this sampling, $Q_{\text{target}}^{(t)}$ converges toward the optimal $Q$ function, $Q^{\star}$. In our implementation, for each episode, the agent was allowed to sample the image for $20$ steps. We set $\gamma = 0.99$, a frequently used value that, being close to $1$, emphasizes current and next states over but includes those further in the future. 

\subsection{Training: Backpropagation of the Deep $Q$ network}

In each step of DQN backpropagation, we randomly select a batch size of $N_{\text{batch}}=128$ transitions from the rows of $\mathbb{T}$ and compute corresponding $Q_{\text{target}}^{(t)}$ and $Q_{\text{CNN}}^{(t)}$ values, yielding the vectors $\vec{Q}_{\text{target}} = \{ Q_{\text{target}}^{(t)} \}_{t=1}^{N_{\text{batch}}}$ and $\vec{Q}_{\text{CNN}} = {Q_{\text{CNN}}^{(t)}} \}_{t=1}^{N_{\text{batch}}}$.  We backpropagate to minimize the loss $L_{\text{batch}}$ of said batch,
\begin{equation}
    L_{\text{batch}} = \frac{1}{N_{\text{batch}}} \sum_{i=0}^{N_{\text{batch}}} \arrowvert Q_{\text{target}}^{(i)} - Q_{\text{DQN}}^{(i)} \arrowvert \text{.}
\end{equation}
Fortunately, as we proceed in training $Q_{\text{DQN}}^{(t)}$ to successively approximate $Q_{\text{target}}^{(t)}$, our CNN function approximation should converge toward that reflecting the optimal policy, so that:
\begin{equation}
    \lim_{t\to\infty} \left( Q_{\text{DQN}}^{(t)} \right) = \lim_{t\to\infty} \left( Q_{\text{target}}^{(t)} \right) = Q^{\star}
\end{equation}

To train the DQN, we employed the Adam optimizer with learning rate of $1 \times 10^{-4}$. We implemented DQN training in the Pytorch package in Python 3.7 executed in Google Colab. 

\begin{figure}

\centering
\advance\leftskip-2cm
\includegraphics[width=14cm,height=10cm]{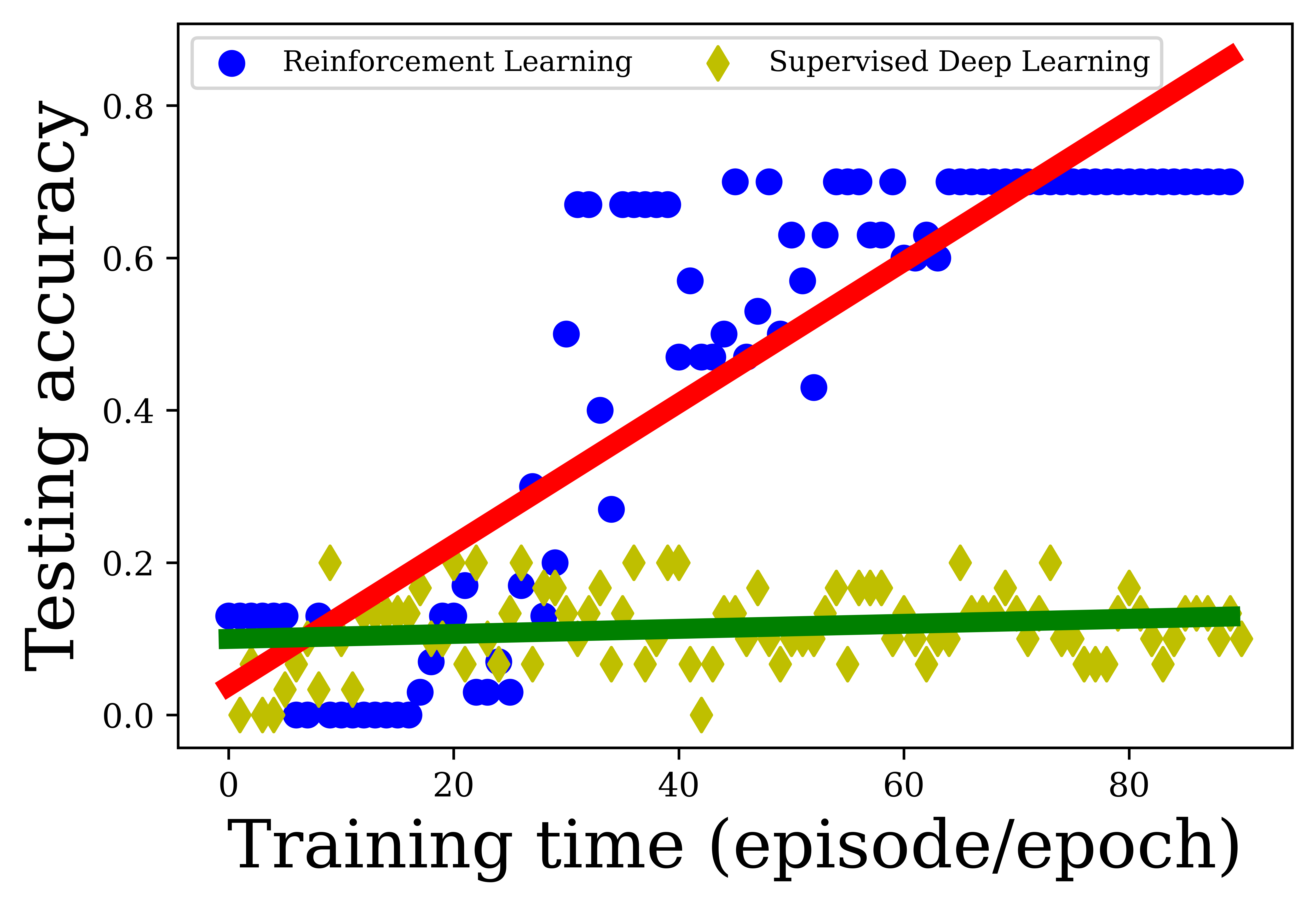}
\caption{Comparison between reinforcement learning (Deep Q learning) and supervised deep learning. The mean accuracy of the testing set of 30 images is plotting as a function of training time, which is measured for reinforcement learning in episodes, and for supervised deep learning in epochs. Both methods were trained on the same 30 training set images. Blue circles represent the results of reinforcement/Deep Q learning, with best fit line in red. Accuracy for supervised deep learning is shown as yellow diamonds, with best fit line in green. Whereas supervised deep learning quickly overfits the small training set and thus cannot learn from it in any generalized manner, the Deep Q network is able to learn over time in a way that can generalize to a separate testing set.}
\label{fig:comp_dqn_vs_keypoint}
\end{figure}

\section*{Results}

We trained the DQN on $30$ two-dimensional image slices from the BraTS database at the level of the lateral ventricles. We did not employ any data augmentation. Training was performed for $90$ episodes. For each of the separate $30$ testing set images, the trained $DQN$ was applied to the initial state with agent in the top left corner and successively to each subsequent state for a total of $20$ steps. 

In order to compare the performance of RL / Deep $Q$ learning with that of standard supervised deep learning, we trained a localization supervised deep learning network as well. More specifically, we trained a keypoint detection CNN with architecture essentially identical to that of our DQN. Again, to make the comparison as fair as possible, we trained the keypoint detection CNN on the same $30$ training images for $90$ epochs. Not surprisingly for such a small training set, the supervised keypoint detection CNN quickly overfit the training set, with training and testing set losses diverging before the $10$th epoch. 

Figure \ref{fig:comp_dqn_vs_keypoint} shows a head-to-head comparison of the two techniques. It plots accuracy of the trained networks on the separate testing sets of 30 images as a function of training time, measured as episodes for DQN and as epochs for supervised deep learning. Supervised deep learning does not learn in a way that generalizes to the testing set, as evidenced by the essentially zero slope of the best fit line during training. DQN learns during training, as manifested by the positive slope of the best fit line. Ultimately, RL/Deep $Q$ learning achieves an average accuracy of $70$\% over the last $20$ episodes, whereas supervised deep learning has a corresponding mean accuracy of $11$\%, a difference that is statistically significant by standard $t$-test, with $p$-value of $5.9 \times 10^{-43}$. 

\section*{Discussion}

We have demonstrated as proof-of-principle that reinforcement learning, here implemented as Deep $Q$ learning, can accurately localize lesions on medical imaging. Specifically, in this demonstration, we have applied the approach to identifying and locating glioblastoma multiforme brain lesions on contrast-enhanced MRI. 

More significantly, in so doing, we have demonstrated that the approach can produce reasonably accurate results with a training set size of merely $30$ images. This number is at least an order of magnitude below what is generally considered necessary for almost any radiology AI application. This follows from the fact that current radiology AI has been dominated by supervised deep learning, an approach that depends on large amounts of annotated data, typically requiring hundreds (or, preferably, thousands) of annotated images to achieve high performance.

To restate the three key limitations of the currently prevalent supervised deep learning in radiology, they are:
1) Requirement of large amounts of expert-annotated data;
2) Susceptibility to grossly incorrect predictions when applied to new data sets; and
3) Lack of insight or intuition into the algorithm.

This proof-of-principle work provides evidence that reinforcement learning can address limitation \#1. It can also address limitation \#3, as evidenced by the reward structure illustrated in Figures \ref{fig:fig_1b}--\ref{fig:fig_1d}. 

Future work will address limitation \#2 by comparing reinforcement learning and supervised deep learning trained on data from one institution and tested on separate images from another institution. 

An important limitation of the present work is that it has been performed on two-dimensional image slices. Future work will extend to fully three-dimensional image volumes. As can be seen in Figure \ref{fig:comp_dqn_vs_keypoint}, the Deep $Q$ learning training process is somewhat noisy. Future work will utilize different techniques in reinforcement learning to learn in a smoother fashion. It should be noted that we tried employing policy-gradient learning  to achieve this less noisy learning. We did so with the actor-critic approach in both its single-agent version, A2C, and its multi-agent form, A3C. Both approaches failed to learn as Deep $Q$ learning could. We suspect that this is caused by the sequential nature of sampling in A2C/A3C, which could not make use of the varied sampling across environments (i.e., different training set images) and states. We anticipate that future work incorporating a replay memory buffer with policy gradient may ultimately work best, and this will be the focus of our future work. 

\section*{Conflicts of interest}

The authors have pursued a provisional patent based largely on the work described here.

\printbibliography

@article{wang2020inconsistent,
  title={Inconsistent Performance of Deep Learning Models on Mammogram Classification},
  author={Wang, Xiaoqin and Liang, Gongbo and Zhang, Yu and Blanton, Hunter and Bessinger, Zachary and Jacobs, Nathan},
  journal={Journal of the American College of Radiology},
  year={2020},
  publisher={Elsevier}
}

@article{goodfellow2014explaining,
  title={Explaining and harnessing adversarial examples},
  author={Goodfellow, Ian J and Shlens, Jonathon and Szegedy, Christian},
  journal={arXiv preprint arXiv:1412.6572},
  year={2014}
}

@article{buhrmester2019analysis,
  title={Analysis of explainers of black box deep neural networks for computer vision: A survey},
  author={Buhrmester, Vanessa and M{\"u}nch, David and Arens, Michael},
  journal={arXiv preprint arXiv:1911.12116},
  year={2019}
}

@article{liu2019comparison,
  title={A comparison of deep learning performance against health-care professionals in detecting diseases from medical imaging: a systematic review and meta-analysis},
  author={Liu, Xiaoxuan and Faes, Livia and Kale, Aditya U and Wagner, Siegfried K and Fu, Dun Jack and Bruynseels, Alice and Mahendiran, Thushika and Moraes, Gabriella and Shamdas, Mohith and Kern, Christoph and others},
  journal={The lancet digital health},
  volume={1},
  number={6},
  pages={e271--e297},
  year={2019},
  publisher={Elsevier}
}

@article{stember2020deep,
  title={Deep reinforcement learning to detect brain lesions on MRI: a proof-of-concept application of reinforcement learning to medical images},
  author={Stember, Joseph and Shalu, Hrithwik},
  journal={arXiv preprint arXiv:2008.02708},
  year={2020}
}

@book{sutton2018reinforcement,
  title={Reinforcement learning: An introduction},
  author={Sutton, Richard S and Barto, Andrew G},
  year={2018},
  publisher={MIT press}
}

@article{menze2014multimodal,
  title={The multimodal brain tumor image segmentation benchmark (BRATS)},
  author={Menze, Bjoern H and Jakab, Andras and Bauer, Stefan and Kalpathy-Cramer, Jayashree and Farahani, Keyvan and Kirby, Justin and Burren, Yuliya and Porz, Nicole and Slotboom, Johannes and Wiest, Roland and others},
  journal={IEEE transactions on medical imaging},
  volume={34},
  number={10},
  pages={1993--2024},
  year={2014},
  publisher={IEEE}
}

\end{document}